\documentclass[conference]{IEEEtran}
\IEEEoverridecommandlockouts

\usepackage{cite}
\usepackage{amsmath,amssymb,amsfonts}
\usepackage{graphicx}
\usepackage{textcomp}
\usepackage{xcolor}
\usepackage[colorlinks=true,linkcolor=red,citecolor=green,urlcolor=blue,]{hyperref}
\usepackage{booktabs}
\usepackage{multirow}
\usepackage{algorithm}
\usepackage{array}
\usepackage{adjustbox}
\usepackage{algpseudocode}

\newcommand{\safeincludegraphics}[2][]{%
  \IfFileExists{#2}{\includegraphics[#1]{#2}}{%
    \fbox{%
      \parbox[b][0.25\textheight][c]{\linewidth}{\centering
        \textbf{Image `#2' not found.}\\[4pt]
        Upload the file to Overleaf or change the filename.}%
    }%
  }%
}

\DeclareGraphicsExtensions{.pdf,.png,.jpg,.jpeg}

\def\BibTeX{{\rm B\kern-.05em{\sc i\kern-.025em b}\kern-.08em
    T\kern-.1667em\lower.7ex\hbox{E}\kern-.125emX}}

\begin{document}

\title{LDP: Parameter-Efficient Fine-Tuning of Multimodal LLM for Medical Report Generation}

% For anonymous submission keep author as below.
% \author{\IEEEauthorblockN{Anonymous ICME submission}}
\author{\IEEEauthorblockN{Tianyu Zhou, Junyi Tang, Zehui Li, Dahong Qian, Suncheng Xiang}}
\maketitle

\begin{abstract}
Colonoscopic polyp diagnosis is pivotal for early colorectal cancer detection, yet traditional automated reporting suffers from inconsistencies and hallucinations due to the scarcity of high-quality multimodal medical data. To bridge this gap, we propose \textbf{LDP}, a novel framework leveraging multimodal large language models (MLLMs) for professional polyp diagnosis report generation. Specifically, we curate \textbf{MMEndo}, a multimodal endoscopic dataset comprising expert-annotated colonoscopy image-text pairs. We fine-tune the Qwen2-VL-7B backbone using Parameter-Efficient Fine-Tuning (LoRA) and align it with clinical standards via Direct Preference Optimization (DPO). Extensive experiments show that our LDP outperforms existing baselines on both automated metrics and rigorous clinical expert evaluations (achieving a Physician Score of 7.2/10), significantly reducing training computational costs by 833× compared to full fine-tuning. The proposed solution offers a scalable, clinically viable path for primary healthcare, with additional validation on the IU-XRay dataset confirming its robustness.
\end{abstract}


\section{Introduction}

Colorectal cancer (CRC) is a leading cause of mortality, with most cases arising from adenomatous polyps~\cite{ref-crc-stats}. While colonoscopy is the gold standard for early detection, its effectiveness relies heavily on the endoscopist’s expertise. Studies report polyp miss rates up to 28\%, highlighting issues of diagnostic inconsistency and inefficiency, particularly in resource-limited primary healthcare settings~\cite{ref-polyp-miss}.

Recent advances in artificial intelligence (AI) have demonstrated strong potential in medical image analysis and automated report generation~\cite{ref-ai-medicine}. Early methods relied on template- or rule-based systems~\cite{ref-template-methods}, while later approaches employed deep learning frameworks such as encoder–decoder architectures~\cite{ref-encoder-decoder} and Transformer-based models~\cite{ref-transformer-medical}. More recently, multimodal large language models (LLMs) have shown remarkable capabilities in integrating vision and language for clinical decision support~\cite{ref-multimodal-llm}. However, in the specific domain of colonoscopy polyp diagnosis report generation, multimodal large models remain unexplored. The primary challenges include maintaining logical coherence across long clinical narratives, ensuring accurate alignment between visual and textual representations, and enabling efficient adaptation under constrained computational budgets~\cite{ref-colon-ai-challenges}.

\begin{figure}[!t]
\centering
% use safe include to avoid missing-file error during compile
\safeincludegraphics[width=1.0\columnwidth]{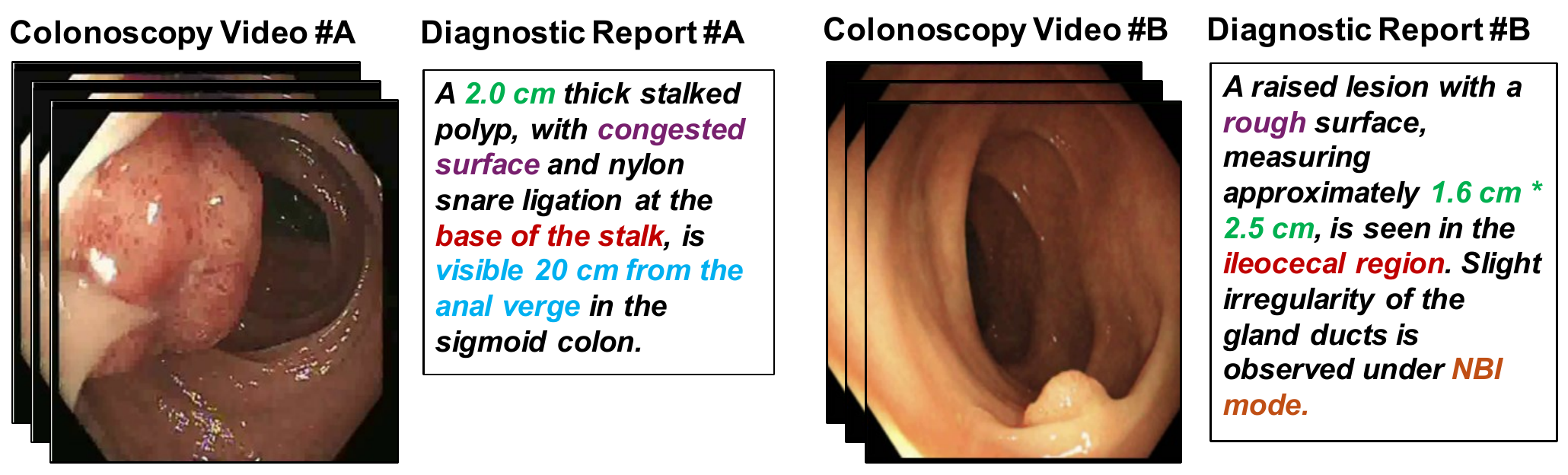}
\caption{Some illustrative visual exemplars sourced from our polyp diagnosis dataset named MMEndo, which encompasses colonoscopy images along with their corresponding diagnostic reports.}
\label{fig-head}
\end{figure}

To address these challenges, we propose a multimodal large-model framework for colonoscopy report generation that integrates visual understanding, textual reasoning, and human preference alignment. The framework leverages the Qwen2-VL-7B backbone and incorporates parameter-efficient fine-tuning techniques such as LoRA~\cite{ref-lora}, as well as Direct Preference Optimization (DPO)~\cite{ref-dpo} to refine model outputs according to clinical preferences. A dedicated multimodal dataset named \textbf{MMEndo} containing colonoscopic images and diagnostic reports was constructed to support domain adaptation and evaluation, shown in Figure~\ref{fig-head}. Rather than pursuing marginal improvements on generic metrics through computationally expensive full-parameter training, we prioritize clinical applicability and efficiency. To this end, we propose the first framework that combines parameter-efficient fine-tuning (LoRA) with preference alignment (DPO) for this specialized medical task, achieving a balance between high performance and deployment feasibility. We demonstrate the feasibility and clinical adaptability of this approach using a high-quality, though small-scale, dataset.

The main contributions of this paper are as follows:
\begin{itemize}
  \item We are the first to integrate multimodal large models, parameter-efficient fine-tuning, and preference alignment into a unified pipeline for medical report generation.
  \item Our approach achieves high performance with minimal computational cost, making it suitable for deployment in resource-limited clinical environments.
  \item Extensive experiments demonstrate that our model achieves superior performance on both automatic assessments and expert-based assessments, ensuring accuracy, coherence, and interpretability of generated reports.
\end{itemize}

Overall, this work provides a feasible technical route for intelligent medical report generation and offers a scalable foundation for extending multimodal large model applications to broader medical imaging and diagnostic scenarios.

\section{Related Works} 

\subsection{Template- and Rule-Based Report Generation}
Early approaches relied on predefined templates and expert rules to map patient data into fixed patterns. Representative methods include the template-based CBCI System~\cite{Sitompul2021}, the retrieval-generation model HRGR-Agent~\cite{Li2018HRGR}, and VTI~\cite{Najdenkoska2021}, which aligns modalities using latent topics. While simple and efficient, these systems are rigid, as fixed slots limit flexibility and adaptation to rare cases. Although later works attempted to incorporate external knowledge, overall adaptability remains constrained.

\subsection{Deep Learning-Based Methods}
With the rise of deep learning, encoder-decoder models using CNNs and RNNs became prevalent. For example, Tiwari et al.~\cite{Tiwari2022} utilized LSTM-CNNs for chest X-ray reports. Subsequently, attention mechanisms and Transformer-based architectures were introduced to better capture long-range context and reduce repetition. While these methods improve content richness over template-based systems, they still face limitations, including inefficiency with long sequences, potential semantic incoherence, and imperfect alignment between visual features and textual descriptions.

\subsection{Large Model-Driven Report Generation}
Recently, multimodal large language models (MLLMs) have shown promise in medical reporting. For instance, Dia-LLaMA~\cite{ChenDiaLLaMA2024} adapts LLaMA2-7B with guidance prompts for CT abnormality detection. Other approaches combine strong image encoders with LLMs to reduce reliance on templates and enhance linguistic quality. However, within the colonic polyp domain, such end-to-end applications are underexplored. Significant challenges remain regarding data privacy, computational costs, and effective domain adaptation for specialized endoscopic scenarios.

\section{Our Method}

\begin{figure}[!t]
\centering
\safeincludegraphics[width=1\columnwidth]{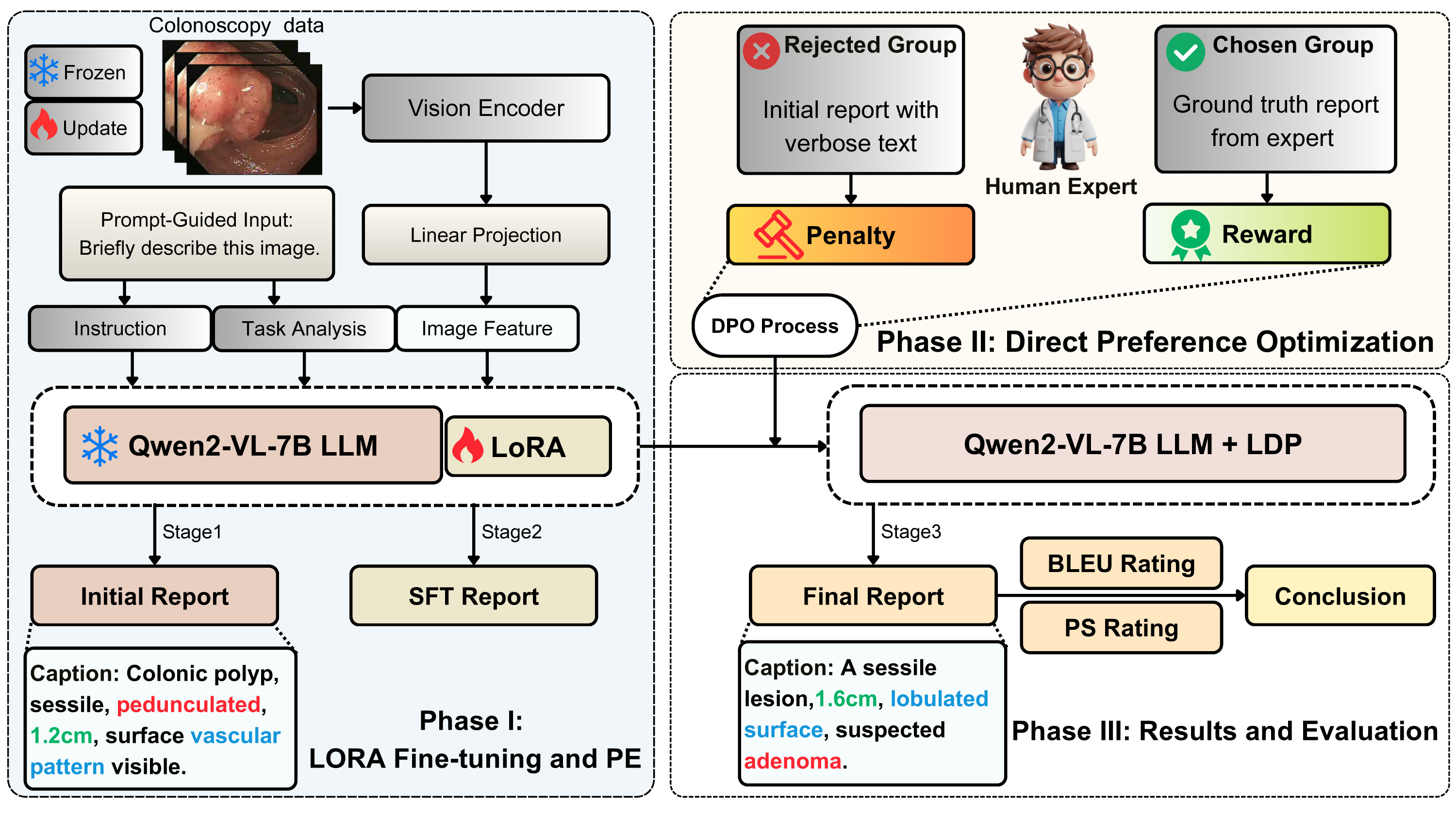}
\caption{Overall framework of the proposed LDP framework, which integrates multimodal large models, parameter-efficient fine-tuning, and preference alignment (DPO) into a unified pipeline for polyp report generation. }
\label{fig-data}
\end{figure}

\subsection{Preliminary}

In this work, we propose an automated algorithm for polyp diagnosis report generation based on multimodal large language models (MLLMs).  We term this framework \textbf{LDP} (\textbf{L}oRA, \textbf{D}PO, and \textbf{P}rompt Engineering), representing a unified pipeline that integrates parameter efficiency, human alignment, and task guidance. Specifically, we utilize the Qwen2-VL-7B pre-trained model and employ a progressive optimization strategy to enhance performance for medical report generation, illustrated in Figure~\ref{fig-data}. First, Multimodal Dataset Construction is performed to achieve crucial domain adaptation. Second, Parameter-Efficient Fine-Tuning (PEFT), specifically LoRA, is employed to efficiently adapt the model under limited computational resources. Finally, Preference Optimization (DPO) is used to align the model's output with clinical quality and human preference, ensuring high accuracy and consistency in the generated reports. 
% We term this novel, efficient, and preference-aligned pipeline the LDP framework, where the acronym stands for LoRA, DPO, and Prompt Engineering (LDP=LoRA+DPO+PE). This integrated approach leverages parameter efficiency (LoRA), human alignment (DPO), and task guidance (PE) to achieve superior clinical performance.

\subsection{MMEndo Dataset Construction}
Our MMEndo dataset comprises 36 colonoscopy videos from 27 patients provided by Zhongshan Hospital of Fudan University with ethical approval. After careful preprocessing and cleaning, we obtained 2,314 image-text pairs for model training and evaluation. The detailed construction and preprocessing steps are as follows.

\textbf{Keyframe Extraction}: The original colonoscopy videos were first segmented to extract clips containing detected polyps. Keyframes were then extracted from these segments using a dynamically adjusted sampling rate based on the video length to capture essential visual information.

\textbf{Data Cleaning}: Specialized annotators filtered out invalid data, discarding images with severe blurring, mucus, or shifted fields. Only clear images containing polyps were retained.

\textbf{Image-Text Alignment}: Diagnostic reports covering core information (e.g., location, size) were written by senior physicians from Zhongshan Hospital with 98.7\% terminology accuracy. We employed a ``frame-to-sentence" alignment strategy to precisely link image regions with textual descriptions, ensuring high-fidelity image-text pairs.

% The final dataset is structured in a standardized JSON format, where each unit includes a query (a detailed instruction for the medical diagnosis task), a response (the professional diagnostic report text) and an images path. 
\subsection{Model Architecture and Fine-tuning}
We select Qwen2-VL-7B as our backbone model, chosen for its state-of-the-art multimodal understanding capabilities and its excellent cost-performance under limited computational resources. It consists of three core components: Vision Encoder (ViT), Language Model (QwenLM Decoder), and Vision-Language Adapter.

Vision Encoder (ViT): The 675M-parameter Vision Transformer supports dynamic resolution via 2D-RoPE to capture spatial information from raw images.

Language Model (QwenLM Decoder): Based on the strong Qwen2 series of Large Language Models, this component is responsible for text generation and multimodal reasoning, utilizing a self-attention mechanism to fuse the combined visual and text inputs.

Vision-Language Adapter (VL Adapter): It utilizes M-ROPE to capture positional information across 1D (text), 2D (image), and 3D (video) modalities. A single-layer cross-attention mechanism compresses visual sequences to address computational efficiency.

For parameter-efficient fine-tuning, we employ LoRA (Low-Rank Adaptation), a Parameter-Efficient Fine-Tuning (PEFT) technique. This is adopted to overcome the high computational cost and resource demands associated with Full Fine-Tuning (FFT). LoRA reduces trainable parameters by decomposing the weight update matrix ($\Delta W$) into two low-rank matrices ($B$ and $A$). The original pre-trained weights ($W$) are frozen, and only the lightweight $B$ and $A$ matrices are optimized, significantly reducing the number of trainable parameters from $d^2$ to $2dr$. We apply LoRA primarily to the weights of the Self-Attention layers (Query, Key, and Value) within the QwenLM Decoder to adapt the model to the polyp diagnosis domain.

\subsection{Preference Optimization}
While fine-tuning with LoRA provides domain adaptation, we further enhance our model using Preference Optimization to align the generated diagnostic reports with human-defined standards for clinical quality and relevance.

We select Direct Preference Optimization (DPO) as the core alignment algorithm. DPO offers a substantial advantage over traditional Reinforcement Learning from Human Feedback (RLHF) methods, such as PPO, because it directly optimizes the policy (model) based on human preference data, thereby eliminating the need to train a separate, explicit reward model.

This approach significantly simplifies the training pipeline and offers greater stability. DPO works by mathematically transforming the optimization objective to maximize the probability of ``preferred" responses relative to ``non-preferred" responses from the collected human preference dataset. This direct optimization allows the model to effectively encode human and clinical preference patterns, resulting in superior output quality and better alignment with clinical expectations. 

To support DPO training, we constructed a specialized clinical preference dataset comprising preferred and non-preferred report pairs. Preferred Samples: We collected authentic diagnostic reports written by expert endoscopists from Zhongshan Hospital and Tongji Hospital. These reports represent the clinical ``gold standard", characterized by their conciseness and high information density.
Non-Preferred Samples: We utilized reports generated by our base model with PE as the non-preferred counterparts. These samples often contain hallucinations or non-standard terminology despite being syntactically correct. By learning from the contrast between these two sets, DPO steers the model to acquire the nuanced style and diagnostic focus of expert physicians, aligning the output towards greater conciseness and professional accuracy.
Furthermore, we also investigate other preference alignment algorithms like SimPO (Simple Preference Optimization) and ORPO (Odds Ratio Preference Optimization) to explore further avenues for model enhancement.
The overall training procedure of LDP method is depicted in Algorithm~\ref{alg:Framwork}.

\begin{algorithm}[!t]
\caption{LDP Training Procedure}
\label{alg:Framwork}
\begin{algorithmic}[1]
\Require Multimodal dataset $\mathcal{D} = \{(I_i, T_i)\}_{i=1}^N$; pre-trained Qwen2-VL-7B model $\theta_{\text{pre}}$; LoRA parameters $\phi$.
\Ensure Optimized model $\theta_{\text{opt}}$.
\State Initialize LoRA parameters $\phi$ for Qwen2-VL-7B
\For{epoch = 1 to E}
\State Sample batch $(I, T) \sim \mathcal{D}$
\State Compute visual features: $F_v = \mathrm{ViT}(I)$
\State Generate text: $\hat{T} = \mathrm{QwenLM}(F_v)$
\State Compute LoRA loss: $\mathcal{L}_{\mathrm{LoRA}} = \mathrm{CE}(T, \hat{T})$
\State Update $\phi \leftarrow \phi - \eta \nabla_{\phi}\mathcal{L}_{\mathrm{LoRA}}$
\EndFor
\State Apply DPO optimization with human preference data
\State \Return $\theta_{\text{opt}}$
\end{algorithmic}
\end{algorithm}

\section{Experiments}

\subsection{Experimental Setup}
% This section explicitly defines the experimental environment, model selection, data processing, and evaluation methodology to ensure the reproducibility of our results.

\textbf{Datasets and Evaluation Metrics.}
We evaluate our proposed method on two datasets. Our MMEndo dataset is a proprietary, single-center collection of colonoscopy images paired with expert diagnostic texts, crucial for specialized medical tasks. The dataset was manually split into a training set and a test set at an 8:2 ratio, employing stratified sampling to ensure the balanced distribution of various polyp types for robust evaluation. To assess the model's out-of-domain robustness, we also employ the \textbf{Generalization Dataset}, the public IU-XRay dataset~\cite{demner2016iu}, applying the LDP framework to the chest X-ray report generation task.

For automated evaluation, we utilize standard natural language generation metrics: \textbf{BLEU-1 to BLEU-4}, \textbf{METEOR}, \textbf{ROUGE-L}, and \textbf{CIDEr}. Additionally, considering the critical nature of medical reports, we introduce a \textbf{Clinical/Qualitative Metric}: the Physician Score (PS). To verify the model's clinical application, we invited five expert physicians from multiple top-tier hospitals in Shanghai (including Tongren Hospital, Ruijin Hospital, and Chest Hospital) to perform a manual evaluation. Specifically, three of these physicians were from Tongren Hospital, and their individual scores were averaged to represent the overall Tongren evaluation, resulting in five distinct score columns in the final table. Reports were scored based on clinical accuracy, factual completeness, and usability (appropriateness of terminology and structure). The PS value is measured on a comprehensive scale of 1 to 10, where 10 represents the highest clinical quality and alignment with professional standards. To validate the reliability of this metric, we calculated the inter-rater reliability using Cohen's Kappa. The resulting $\kappa$ coefficient was 0.72 (95\% CI: 0.65--0.79), indicating a high degree of agreement (Substantial Agreement) among the seven expert evaluators and confirming the objectivity of the PS metric.

\textbf{Implementation Details.}
We implement our approach using PyTorch, training on 4$\times$ NVIDIA RTX 4090 GPUs. The \textbf{Base Model} used is the multimodal large model Qwen2-VL-7B. For the Supervised Fine-Tuning (SFT) phase, we set the learning rate to $2\times10^{-4}$ and use a batch size of 16. The \textbf{PEFT Parameters} utilize LoRA with a rank r and scaling factor $\alpha$. Specifically, we apply LoRA modules to the attention projection matrices (Q, K, V, O) within the Vision-Language fusion layers and the Language Model decoder blocks. The \textbf{DPO Optimization} phase uses a learning rate of $1\times10^{-6}$ with a preference weight $\beta$ of 0.1, focusing on aligning the model output with expert-preferred reports constructed from the SFT dataset.

\subsection{Comparison with SOTAs}
This section demonstrates the superior performance of the LDP framework on medical report generation compared to state-of-the-art (SOTA) methods.

\textbf{Quantitative Analysis.}
Table~\ref{tab1} presents the performance of our proposed LDP method against several strong baselines on our proprietary polyp dataset. The baselines include the vanilla Qwen2-VL-7B model, models enhanced with simple parameter initialization (PE), and other competitive PEFT strategies like AdaLoRA.

\begin{table}[!t]
\centering
\caption{Performance comparison on our MMEndo dataset (\textbf{Bold} indicates best performance). Note: PS is measured on a 1--10 scale.}
\small
\setlength{\tabcolsep}{0.7mm}{
\begin{tabular}{p{20mm}cccccc}
\toprule
Method & BLEU-1 & BLEU-2 & METEOR & ROUGE-L & CIDEr & PS \\
\midrule
Qwen2-VL-7B & 0.123 & 0.041 & 0.107 & 0.074 & 0.098 & 3.2 \\
+ PE & 0.182 & 0.072 & 0.157 & 0.117 & 0.213 & 4.5 \\
+ AdaLoRA & 0.374 & 0.243 & 0.296 & 0.347 & 0.402 & 5.8 \\
+ LoRA (SFT) & 0.642 & 0.585 & 0.609 & 0.638 & 0.527 & 6.7 \\
LDP (Ours) & \textbf{0.658} & \textbf{0.591} & \textbf{0.618} & \textbf{0.636} & \textbf{0.520} & \textbf{7.2} \\
\bottomrule
\end{tabular}}
\label{tab1}
\end{table}

LoRA achieved a 5.2x increase in BLEU-1 (from 0.123 to 0.642), and DPO further boosted PS score to 7.2, validating the synergistic value of ``Parameter-Efficient Fine-Tuning + Preference Alignment".

The results show that the full LDP framework (\textbf{LDP (LoRA+DPO+PE)}) achieves the best overall performance, particularly demonstrating a significant jump in the specialized \textbf{PS} metric (from 6.7 to 7.2 on a 10-point scale), underscoring the effectiveness of the DPO-based preference alignment in generating clinically superior reports.

\textbf{Qualitative Analysis.}
The \textbf{PS} column in Table~\ref{tab1} reflects the average results of the expert physician evaluation. The LDP method reaches an average score of 7.2 out of 10, rated as ``Good" by the experts. Table~\ref{tab4} provides the detailed breakdown of the Physician Scores across the seven invited specialists. The results confirm that integrating preference optimization successfully aligns the model's output with professional standards, which is often difficult to capture with purely automated metrics. The overall positive score indicates the model's strong practical utility and reference value in clinical applications.
To ensure consistency and objectivity in our manual evaluation, all seven participating physicians were trained using the standardized scoring rubric detailed in Table~\ref{tab:rubric}. Scores were assigned independently. The final PS score was calculated as the average after discarding the single highest and lowest scores to reduce outlier bias.

\begin{table}[!t]
\centering
\caption{Manual Physician Score (PS) Results (Average Score on 4 Selected Cases, Max Score=10). Note: The Tongren score is an average from three individual expert evaluations.}
\small
\setlength{\tabcolsep}{1.05mm}{
\begin{tabular}{p{8mm}cccccc}
\toprule
Model & Tongren & Ruijin\#A & Ruijin\#B  & Ruijin\#C  & Thoracic & \textbf{Average} \\
\midrule
LDP  & 6.0 & 8.5 & 6.5  & 7.0  & 8.0 & \textbf{7.2} \\
\bottomrule
\end{tabular}}
\label{tab4}
\end{table}

\begin{table}[!t]
\centering
\caption{Physician Score (PS) Standardized Scoring Rubric}
\small
\begin{adjustbox}{width=\columnwidth}
\begin{tabular}{lllc}
\toprule
\textbf{Dimension} & \textbf{1 Point (Lowest)} & \textbf{10 Points (Highest)} & \textbf{Weight} \\
\midrule
Clinical Accuracy & Core features (location/size) are wrong. & All core features are 100\% accurate. & 40\% \\
Factual Completeness & Misses $\geq$3 core diagnostic items. & Covers all core items; no omissions. & 30\% \\
Terminology & $\geq$3 errors or non-standard terms. & Terminology is precise and standard. & 20\% \\
Clinical Usability & Report is misleading / no reference value. & Report can be directly used for diagnosis. & 10\% \\
\bottomrule
\end{tabular}
\end{adjustbox}
\label{tab:rubric}
\end{table}

\subsection{Ablation Studies}
To analyze the contribution of each proposed component and validate the design choices of the LDP framework, we conduct a detailed ablation study on our MMEndo dataset, summarized in Table~\ref{tab3}.

\begin{table}[!t]
\centering
\caption{Ablation Study on Core Components of LDP Framework (PS scale 1--10)}
\small
\setlength{\tabcolsep}{0.3mm}{
\begin{tabular}{lcccccc}
\toprule
Method & BLEU-1 & BLEU-4 & METEOR & ROUGE-L & CIDEr & PS \\
\midrule
Qwen2-VL-7B & 0.123 & 0.005 & 0.107 & 0.074 & 0.098 & 3.2 \\
 + PE & 0.182 & 0.009 & 0.157 & 0.117 & 0.213 & 4.5 \\
 + LoRA (SFT) & 0.642 & 0.527 & 0.609 & 0.638 & 0.527 & 6.7 \\
 + LoRA + SimPO & 0.645 & 0.531 & 0.612 & 0.630 & 0.524 & 6.7 \\
 + LoRA + ORPO & 0.640 & 0.522 & 0.605 & 0.633 & 0.518 & 6.6 \\
% \midrule
LDP (Ours) & \textbf{0.658} & \textbf{0.545} & \textbf{0.618} & \textbf{0.636} & \textbf{0.520} & \textbf{7.2} \\
\bottomrule
\end{tabular}}
\label{tab3}
\end{table}

\textbf{Contribution of SFT Components: PE and LoRA.}
We first evaluate the necessity of individual components within the Supervised Fine-Tuning (SFT) phase against the zero-shot Baseline.
The introduction of Prompt Engineering (PE) alone significantly improves performance (e.g., BLEU-1 from 0.123 to 0.182, METEOR from 0.107 to 0.157). This confirms PE's essential role in guiding the model towards the required report structure and enhancing the output's adherence to expert formatting.
The combination of Baseline + LoRA (SFT) yields the most substantial performance leap (BLEU-1 reaches 0.642, a 5.2$\times$ increase over the Baseline), validating the critical role of LoRA in adapting the pre-trained Qwen2-VL-7B model to the small-sample, specialized medical task.
Furthermore, we investigated the adaptive PEFT alternative, AdaLoRA. In our experiments, AdaLoRA led to unstable training and a substantial performance drop (BLEU-1 of 0.374), confirming that in this specialized, data-limited scenario, the static, targeted fine-tuning provided by LoRA is superior to AdaLoRA's dynamic rank adjustment.

\begin{figure}[t]
    \centering
    \includegraphics[width=0.50\linewidth]{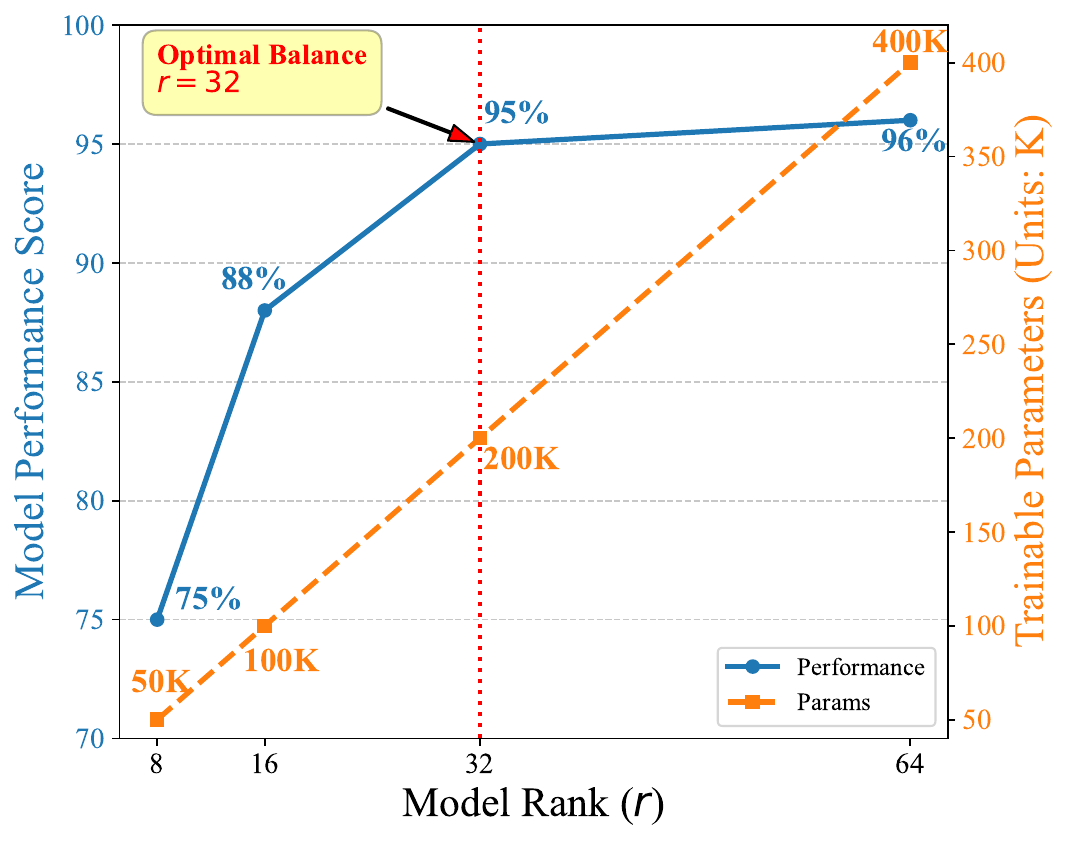}%
    \hfill
    \includegraphics[width=0.50\linewidth]{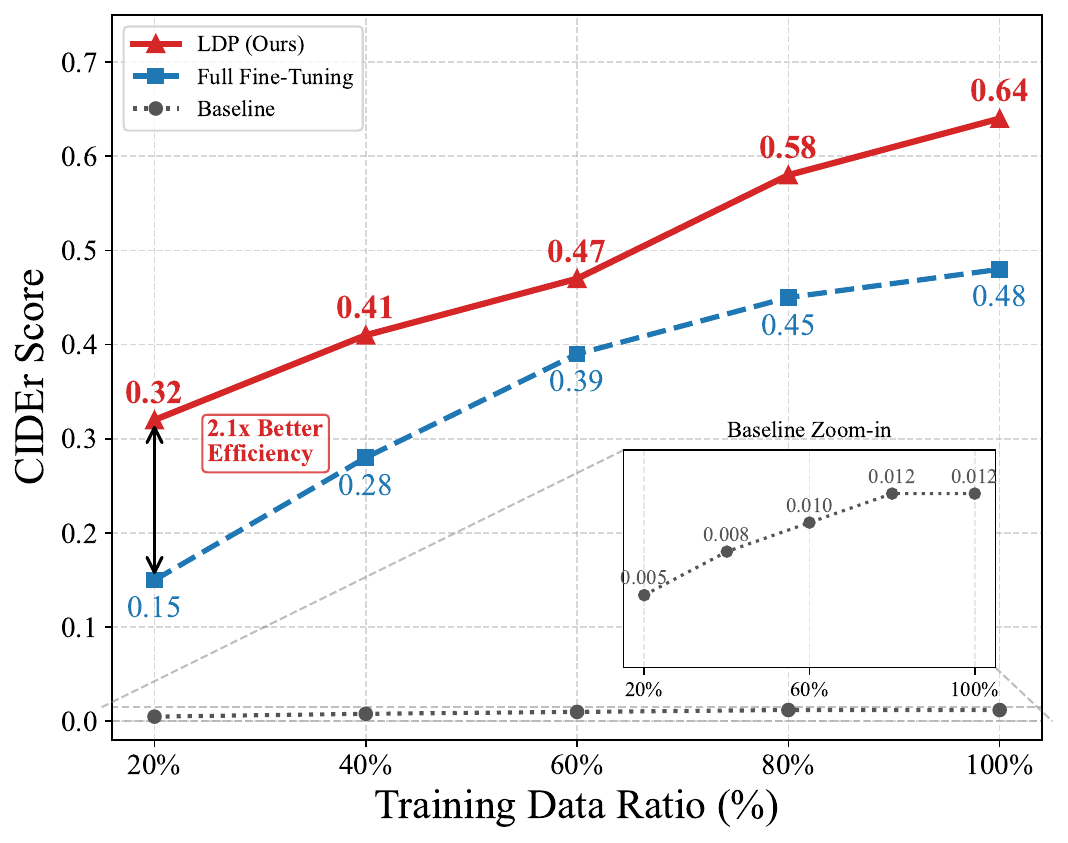}
    
    % 图注里手动说明 Left 和 Right
    \caption{\textbf{Ablation Analysis.} (Left) Impact of LoRA rank (r): r=32 is optimal. (Right) Data efficiency: LDP outperforms baselines, showing 2.1x gain with 20\% data. Zoom in for best view.}
    \label{fig:ablation}
\end{figure}

\textbf{Comparison of Preference Alignment Algorithms.}
We then isolate the impact of the final preference alignment stage, comparing the SFT-optimized model (\textbf{Baseline + LoRA (SFT)}) with the three optimization algorithms. The complete LDP framework (incorporating DPO) achieves the highest Physician Score (PS=7.2), demonstrating that DPO effectively enhances clinical professionalism, fluency, and semantic consistency without sacrificing general performance (BLEU-2/3 metrics also show marginal improvement over SFT-only).
In contrast, SimPO and ORPO showed limited or detrimental effects (PS = 6.7 and 6.6, respectively, and other metrics were dropped in extended analysis):
\begin{itemize}
    \item \textbf{SimPO:} This method employs length normalization to reduce verbosity. However, medical reports require detailed, descriptive text for factual completeness. This mechanism caused SimPO to \textbf{suppress necessary clinical details} (evidenced by the observed drop in CIDER and ROUGE-L in our detailed analysis), making it ill-suited for this task.
    \item \textbf{ORPO:} Designed for single-stage SFT and alignment. In our multi-stage LDP pipeline, where a robust SFT base is established by LoRA, ORPO was \textbf{less stable and effective} than DPO at fine-tuning the subtle preference boundary. Our direct test of ORPO on the zero-shot baseline (without the LoRA SFT base) resulted in severely degraded performance (BLEU-1 of 0.469), confirming that ORPO requires the foundational task-specific adaptation provided by the preceding SFT step.
\end{itemize}
DPO proved to be the most effective algorithm, capable of refining the SFT policy model using explicit preference loss, leading to the highest clinically validated results.

\subsubsection{Analysis of LoRA Rank and Efficiency}
In this section, we investigate the impact of rank $r$ on model performance (BLEU-4, ROUGE-L) and parameter efficiency, and perform experiments with different settings, such as $r=8, 16, 32, 64$. According to \textbf{Figure~\ref{fig:ablation} (Left)}, it can be easily observed that an optimal rank (e.g., $r=32$) provides the best balance between model performance and calculation parameters. Higher ranks lead to diminishing returns in performance but disproportionately increase computational costs, confirming the high parameter efficiency of our PEFT strategy.

\subsection{Generalization and Efficiency Analysis}

\textbf{Generalization Validation.}
To demonstrate that LDP is a general-purpose framework beyond colonoscopy, we extended our evaluation to the IU-XRay dataset. This zero-shot validation confirms our method's ability to adapt MLLMs to diverse medical domains. Table~\ref{tab2} compares LDP against established SOTA methods for medical report generation on this dataset.

\begin{table}[!t]
\centering
\caption{Performance comparison on IU-XRay dataset (\textbf{Bold} indicates best performance)}
\small
\setlength{\tabcolsep}{2.3mm}{
\begin{tabular}{lcccc}
\toprule
Methods & BLEU-1 & METEOR & ROUGE-L & CIDEr \\
\midrule
SaT~\cite{Xu2015ShowAA} & 0.295 & 0.162 & 0.307 & 0.285 \\
AAtt~\cite{Lu2017AdaptiveAC} & 0.312 & 0.173 & 0.321 & 0.297 \\
Transformer~\cite{Vaswani2017AttentionIA} & 0.334 & 0.186 & 0.339 & 0.312 \\
R2GEN~\cite{Chen2020R2GenME} & 0.356 & 0.204 & 0.358 & 0.331 \\
PPKED~\cite{Wang2018TieredAA} & 0.378 & 0.216 & \textbf{0.372} & \textbf{0.351} \\
LDP (Ours) & \textbf{0.392} & \textbf{0.228} & 0.344 & 0.333 \\
\bottomrule
\end{tabular}}
\label{tab2}
\end{table}

The superior performance of LDP, particularly achieving the highest METEOR score, confirms its strong cross-domain \textbf{generalization capability}. This suggests the parameter-efficient and preference-aligned fine-tuning of Qwen2-VL-7B results in robust, transferable visual-language features for broader medical imaging tasks.

\begin{table}[!t]
\centering
\caption{Efficiency Comparison: LDP vs. Full Fine-Tuning (FFT)}
\small
\begin{adjustbox}{width=\columnwidth}
\begin{tabular}{lccc}
\toprule
\textbf{Metric} & \textbf{LDP (Ours)} & \textbf{FFT (Qwen2-VL 7B)} & \textbf{Advantage} \\
\midrule
Trainable Params & 8.4M (0.12\%) & 7.0B (100\%) & 833$\times$ Reduction \\
Training Time & 1.8 Hours & $\sim$48 Hours (Est.) & 27$\times$ Speedup \\
GPU VRAM (Train) & 24 GB (1x 4090) & $\sim$120 GB (2x A100) & 5$\times$ Reduction \\
\bottomrule
\end{tabular}
\end{adjustbox}
\label{tab:efficiency}
\end{table}

A critical advantage of the LDP framework is its computational efficiency, as quantified in Table~\ref{tab:efficiency}. We compare our LDP method (using LoRA with rank r=32) against a theoretical Full Fine-Tuning (FFT) baseline on our 4$\times$ NVIDIA RTX 4090 GPUs.

\begin{itemize}
    \item \textbf{Trainable Parameters:} LDP only requires updating 8.4 million parameters (0.12\% of the total 7B parameters), by applying LoRA to attention layers. This is an 833-fold reduction compared to FFT.
    \item \textbf{Training and Hardware:} LDP completed SFT and DPO phases in 1.8 hours and fits on a single 24GB RTX 4090. In contrast, FFT is estimated to require $\sim$48 hours and $\sim$120GB of VRAM, making it impractical for most clinical settings.
    \item \textbf{Deployment Feasibility:} By maintaining the original architecture, the model can be easily quantized or served via standard efficient inference backends (e.g., vLLM), supporting real-time clinical requirements.
\end{itemize}
This analysis confirms LDP provides an essential, practical solution for deploying advanced LLMs in clinical settings with limited computational resources.

\subsection{Qualitative Results and Case Study}
\begin{figure}[!t]
\centering
\safeincludegraphics[width=1.00\columnwidth]{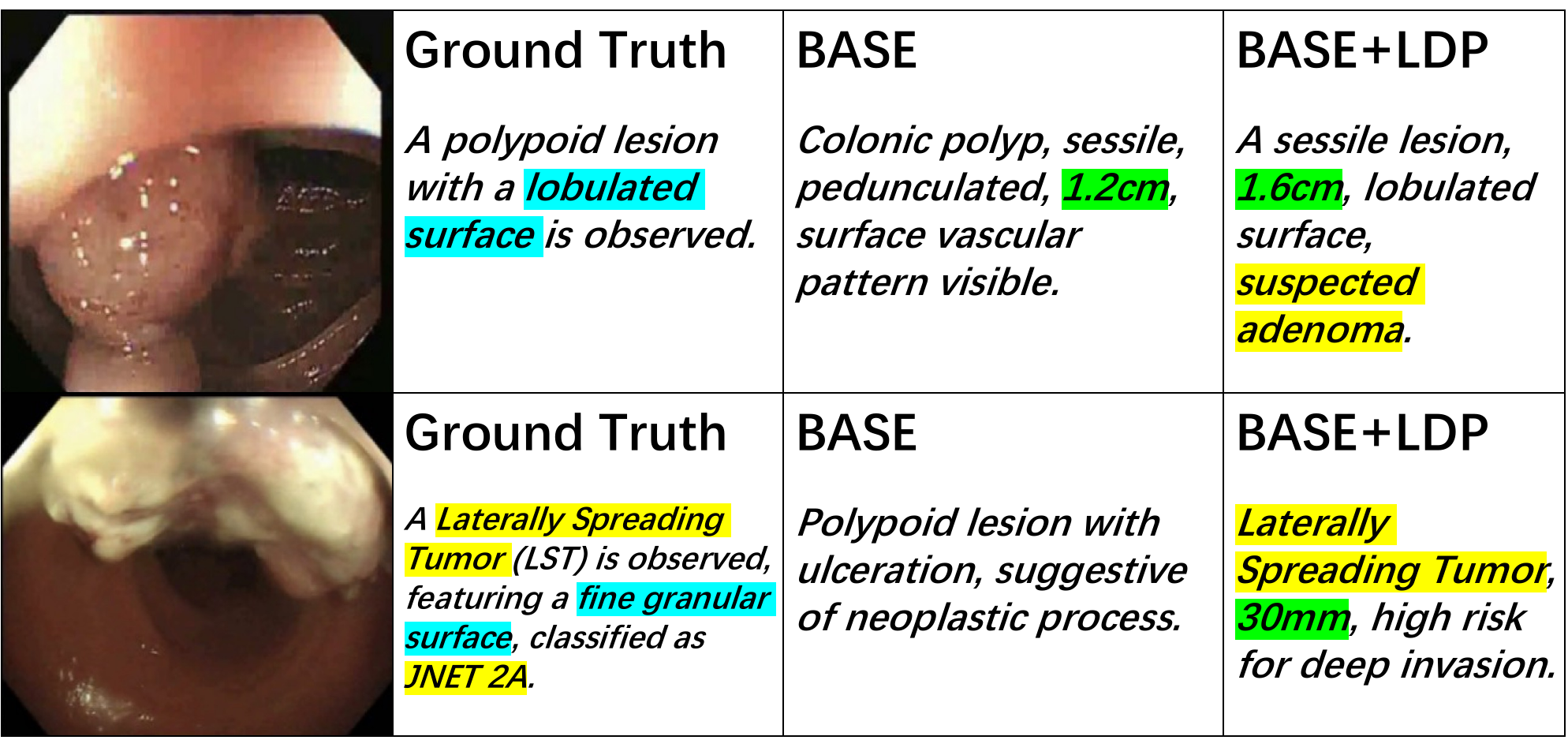}
\caption{Illustrations of reports from ground-truth, BASE+PE and BASE+LDP models for two polyp images. To better distinguish the content in the reports, different colors highlight different medical terms.}
\label{fig:image_in_figure}
\end{figure}

We present qualitative results to offer a visual and textual comparison of the generated reports. Figure~\ref{fig:image_in_figure} illustrates three typical colonoscopy cases. For each case, we show the original image, the expert-provided Ground Truth report, the report generated by the LDP framework, and a leading SOTA baseline (e.g., LoRA SFT-only).

The qualitative analysis demonstrates LDP's ability to not only identify key features but also to structure the output with appropriate terminology and organization, which is a direct outcome of the DPO alignment.

\subsection{Limitations}
The proposed method in this paper can achieve accurate visual and textual representations for medical report generation. Despite its promising performance on this task, we note that there are several limitations in this work. First, our method relies heavily on the specific base model. Consequently, the performance of our LDP is constrained by the Qwen2-VL 7B base model, which requires re-validation of fine-tuning strategies on other multimodal models. Second, DPO primarily teaches expert style and tone of target domain, but it cannot fundamentally alter the model's visual comprehension, resulting in limited PS score improvement. 
These challenges warrant further
research and consideration when deploying LDP model in real scenarios.

\section{Conclusion}
We have presented a novel framework for automated polyp diagnosis report generation that integrates multimodal large models with parameter-efficient fine-tuning and preference optimization. Our approach demonstrates excellent performance while maintaining computational efficiency, making it suitable for deployment in resource-constrained healthcare settings. Extensive experiments on both private and public datasets confirm the effectiveness and generalization capability of our method. Future work will focus on expanding the dataset size. While the current MMEndo dataset is relatively small compared to general domain datasets, it represents a high-quality, expert-verified resource in a specialized medical domain where data privacy severely restricts availability. We aim to explore knowledge-enhanced generation techniques to further mitigate data scarcity.

% ---------------- bibliography ----------------
\bibliographystyle{IEEEtran}
\bibliography{myref}

\end{document}